\documentclass[runningheads]{llncs}
\usepackage{amsmath}
\usepackage{graphicx}
\usepackage{cite}
\usepackage{amsfonts}
\usepackage[margin=1.35in]{geometry}

\newcommand\blfootnote[1]{%
  \begingroup
  \renewcommand\thefootnote{}\footnote{#1}%
  \addtocounter{footnote}{-1}%
  \endgroup
}

\begin{document}

\title{Temporal Factorization of\\3D Convolutional Kernels}
%
%
\author{Gabri\"{e}lle Ras\and
Luca Ambrogioni\and
Umut G\"{u}\c{c}l\"{u}\and
Marcel A. J. van Gerven}
\authorrunning{G. Ras et al.}
%
\institute{Department of Artificial Intelligence, Radboud University, \\Donders Centre for Cognition, Nijmegen, the Netherlands
\email{\{g.ras,l.ambrogioni,u.guclu,marcel.vangerven\}@donders.ru.nl}}

\maketitle              
\begin{abstract} 3D convolutional neural networks are difficult to train because they are parameter-expensive and data-hungry. To solve these problems we propose a simple technique for learning 3D convolutional kernels efficiently requiring less training data. We achieve this by factorizing the 3D kernel along the temporal dimension, reducing the number of parameters and making training from data more efficient. Additionally we introduce a novel dataset called Video-MNIST to demonstrate the performance of our method. Our method significantly outperforms the conventional 3D convolution in the low data regime ($1$ to $5$ videos per class). Finally, our model achieves competitive results in the high data regime ($>10$ videos per class) using up to $45\%$ fewer parameters. 
\blfootnote{Originally published in the proceedings of BNAIC/BENELEARN 2019  CEUR-WS.org/Vol-2491/short106.pdf}
\blfootnote{Copyright © 2019 for this paper by its authors. Use permitted under Creative Commons License Attribution 4.0 International (CC BY 4.0).}

\end{abstract}
\section{Introduction}
Modern deep learning has celebrated tremendous success in the area of automatic feature extraction from data with a grid-like structure, such as images. This success can be largely attributed to the convolutional neural network architecture~\cite{lecun1989generalization}, specifically 2D convolutional neural networks (CNNs).
These networks are successful due the principles of sparse connectivity, parameter sharing and invariance to translation in the input space~\cite{goodfellow2016deep}. Loosely said, 2D CNNs efficiently find class-discriminating local features independent of where they appear in the input space. 
Since video is essentially a sequence of images/frames, 
2D CNNs can be and are used to extract features from the individual frames of the sequence~\cite{karpathy2014large}.
However, the drawback of this method is that the temporal information between frames is discarded. 
Temporal information is important when we want to perform tasks on video such as gesture, action and emotion recognition or classification.
One possible way to simulate the use of time is to stack a recurrent layer after the convolutional layers~\cite{yue2015beyond}.
But correlated spatiotemporal features will not be learnt because spatial and temporal features are explicitly learned in separate regions of the network.
To solve this problem~\cite{baccouche2011sequential} proposed to expand the 2D convolution into a 3D convolution, essentially treating time as a third dimension. Ref.~\cite{ji20123d} used these 3D convolutions to build a 3D CNN for action recognition without using any recurrent layers. 
It is important to notice that the principles that govern 2D CNNs also govern 3D CNNs. Translation invariance in time is useful because the precise beginning and ending of an action are typically ill-defined~\cite{varol2017long}. 
Even though 3D CNNs have been shown to work for different kinds of tasks on video data, they remain difficult to train. 
There are roughly three main issues with 3D CNNs. First, they are parameter-expensive, requiring an abundance of GPU memory. Second, they are data-hungry, requiring much more training data compared to their 2D counterparts. And third, the increase in free parameters leads to a larger search space. As a result these models can be unstable and take a longer time to train.
Existing literature tries to solve these problems by essentially avoiding the use of 3D convolutions completely. The most common method is the factorization of the 3D convolution into a 2D convolution followed by a 1D convolution at the layer level~\cite{qiu2017learning, tran2018closer} or at the network level~\cite{sun2015human, lea2016temporal, lea2016segmental}.

\subsection{Contribution}
We propose a simple and novel method to structure the way 3D kernels are learned during training. This method is based on the idea that nearby frames change very little in appearance. Each 3D convolutional kernel is represented as one 2D kernel with a set of transformation parameters. The 3D kernel is then constructed by sequentially applying a spatial transformation~\cite{jaderberg2015spatial} \textit{directly inside the kernel}, allowing spatial manipulation of the 2D kernel values. We achieve the following benefits:
\begin{itemize}
    \item A reduction in the size of the search space by imposing a sequential prior on the kernel values;
    \item a reduction in the number of parameters in the 3D convolutional kernel;
    \item efficient learning from fewer videos.
\end{itemize}

\section{Related Work}


Previously, an entire 3D convolutional neural network was factorized into separate spatial and temporal layers called factorized spatio-temporal convolutional networks~\cite{sun2015human}. This was achieved by decomposing a stack of 3D convolutional layers into a stack of spatial 2D convolutional layers followed by a temporal 1D convolutional layer. 
Ref.~\cite{tran2018closer} followed in this line of research by factorizing the individual 3D convolutional filters into separate spatial and temporal components called R(2+1)D blocks.
Both methods managed to separate the temporal component from the spatial one. One on the network level~\cite{sun2015human} and one on the layer level~\cite{tran2018closer}. To our knowledge our approach provides the first instance of a temporal factorization at the single kernel level. In effect, we applied the concept of the spatial transformer network~\cite{jaderberg2015spatial} to the 3D convolutional kernel to obtain a factorization along the temporal dimension.

\section{Methods}
The proposed method uses fewer parameters compared to regular 3D convolutions and it imposes a strong sequential dependency on the relationship between temporal kernel slices. In theory our method should allow efficient feature extraction from video data, using fewer parameters and fewer data. The method is explained in Section \ref{sec:method}. We demonstrate the performance of our method on a variant of the classic MNIST dataset~\cite{lecun1998gradient} which we call Video-MNIST. The details of this dataset are explained in Section \ref{sec:video_mnist}. As models we implement 3D and 3DTTN variants of LeNet-5: LeNet-5-3D and LeNet-5-3DTTN respectively (see Section~\ref{sec:models}). Training and inference details are explained in Section~\ref{sec:training}.

\subsection{Temporal factorization of the 3D convolutional kernel}
\label{sec:method}

\begin{figure}
    \centering
    \includegraphics[width=0.75\linewidth]{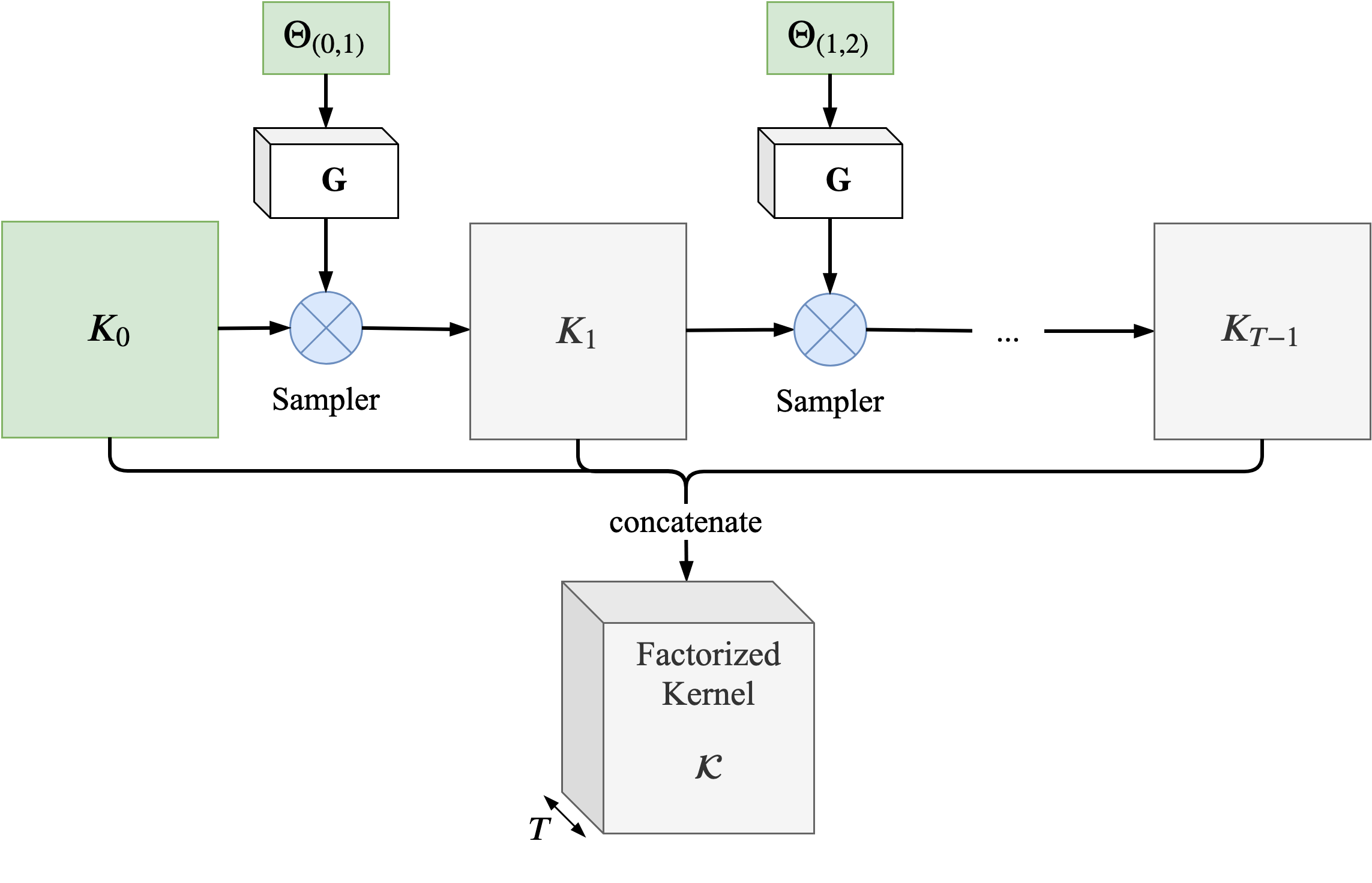}
    \caption{The architecture of the temporally factorized 3D kernel. Each temporal kernel slice is sampled using the previous temporal slice and a transformation matrix. The resulting kernel slices are concatenated to form on the 3D kernel. Instead of learning the entire kernel we only learn $K_0$ and the set of transformations $\mathbf{\Theta}$.}
    \label{fig:diagram}
\end{figure}

\noindent Consider a 3D convolutional layer consisting of $N$ 3D kernels. 
We focus on the inner workings of a single kernel $\mathcal{K} \in \mathbb{R}^{T \times W \times H}$, where $T$, $W$ and $H$ refer to the temporal resolution, width and height of the kernel respectively. 
Without loss of generality, we will assume that the input has a channel with a dimension of one. 


If we slice $ \mathcal{K}$ along the temporal dimension, we end up with $T$ 2D kernels $K \in \mathbb{R}^{W \times H}$. Let us refer to the temporal slice at $t=0$ as $K_0$. Instead of learning entire $ \mathcal{K}$ directly, we only learn $K_0$ and $ \mathbf{\Theta} \in \mathbb{R}^{(T-1) \times 2 \times 3}$. We factorize $ \mathcal{K}$ such that $K_t$ with $t>0$ depends indirectly on $K_0$ via $K_{t+1} = f(K_{t}; \Theta_{(t,t+1)})$ where $\Theta_{(t, t+1)} \in \mathbb{R}^{3 \times 2}$ with $1 \leq t \leq T-1$ are the learnable parameters of the transformation function $f$. For every pair of slices $(K_t, K_{t+1})$ we have
\begin{equation}
    \Theta_{(t, t+1)} = 
    \begin{bmatrix} 
        \theta_{11}  & \theta_{12}   & \theta_{13} \\
        \theta_{21}  & \theta_{22}   & \theta_{23}
    \end{bmatrix}    
\end{equation}

\noindent $\mathbf{\Theta}$ can be further restricted to only contain affine transformation parameters. That is, scaling $s$, rotation $r$, translation in the horizontal direction $t_x$ and translation in the vertical direction $t_y$. This yields:
\begin{equation}
    \Theta_{(t, t+1)} = 
    \begin{bmatrix}
        s \cos r, &-s\sin r, &t_x s\cos r -t_y s\sin r\\
        s\sin r, &s\cos r, &t_x s\sin r +t_y s\cos r 
    \end{bmatrix}
\end{equation}
In that case, $\mathcal{K}$ has only $W \cdot H + 4(T - 1)$ free parameters. We can additionally add the restriction that there is only one shared transformation per kernel. That is, $\Theta_{(t, t+1)} = \Theta_{(t+1, t+2)} = ... =\Theta_{(T-1, T)}$ for $0 \leq t \leq T-1$. This results in just $W \cdot H + 4$ parameters.
Essentially, given $\Theta$, $f$ modifies $K_{t}$ to become $K_{t+1}$, sequentially building the 3D kernel from $K_{0}$. This way we impose a strong sequential relationship between the slices along the temporal dimension of our kernel. 

The nonlinear transformation $f(K; \Theta)$ is applied in two stages. First, $\Theta$ is transformed into a sampling grid $\mathbf{G}$ that matches the shape of the input feature map, plus an explicit dimension for each spatial dimension, $\mathbf{G} \in \mathbb{R}^{W \times H \times 2}$.
Here $K_{t}$ is the input feature map and $K_{t+1}$ is the output feature map. 
We should think of this $\Theta \mapsto \mathbf{G}$ transformation as an explicit spatial mapping of $\Theta$ into the input feature space. 
Each coordinate $(x, y)$ from the input space is split in separate $G_x$ with $x \in [1,\ldots, W]$ and $G_y$ with $y \in [1, \ldots, H]$ components, and calculated as
\begin{equation}
    \begin{bmatrix} 
        G_x\\
        G_y
    \end{bmatrix}
=
    \begin{bmatrix} 
        \theta_{11}  & \theta_{12}   & \theta_{13} \\
        \theta_{21}  & \theta_{22}   & \theta_{23}
    \end{bmatrix} 
    \begin{bmatrix} 
        x\\
        y\\
        1
    \end{bmatrix}
\end{equation}

Now that we have sampling grid $\mathbf{G}$ we can obtain a spatially transformed output feature map $K_{t+1}$ from our input feature map $K_{t}$. To interpolate the values of our new temporal kernel slice we use bilinear interpolation. For one particular pixel coordinate $(x, y)$ in the output map we compute
\begin{equation}
    K_{t+1, x, y} = \sum_{i=1}^h \sum_{j=1}^w K_{t, i, j} \max(0, 1-|G_x - i|) \max(0, 1-|G_y - j|) \,.
\end{equation}

Given that our method transforms temporal kernel slices, we refer to 3D kernels composed with our method as 3DTT kernels. Convolutional networks that use 3DTT kernels instead of regular 3D kernels are referred to as 3DTT convolutional networks or 3DTTNs.

\subsection{Video-MNIST}
\label{sec:video_mnist}
\begin{figure}
    \centering
    \includegraphics[width=0.90\linewidth]{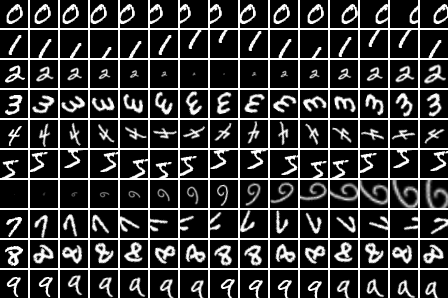}
    \caption{One example from each Video-MNIST class. Instead of displaying the full 30 frames sequence we display 15 frames, skipping one frame each time, such that the figure fits within the margins of the page. 
    }
    \label{fig:video_mnist}
\end{figure}

In order to test our method we constructed a dataset, referred to as Video-MNIST, in which each class has a different appearance and dynamic behavior. Video-MNIST is a novel variant of the popular MNIST dataset. It contains 70000 sequences, each sequence containing 30 frames showing an affine transformation on a single original digit moving in a $28 \times 28$ pixel frame. The class-specific affine transformations are restricted to scale, rotation and x, y translations; see Table~\ref{tab:class_transformations}. We maintain the same train-validation-test split as in the original MNIST dataset. To make the problem more difficult and reliant on both spatial and motion cues, classes $0$, $1$, $5$ and $7$ and $9$ contain random variations of their specific transformation respectively. For classes $0$, $1$, and $7$ the initial direction (left or right) and the initial velocity at which the digit travels per frame is varied. In class $5$ the direction of rotation and the size of the radius of the circular path are varied. In addition we also allow the digits to go partially out of frame or almost vanish ($2$ and $6$). We also made sure that there are overlapping movements between classes, such as rotation or translation in the same direction ($3$, $4$ and $8$). Finally some classes can appear visually similar because of the transformation ($6$ and $9$). In Figure \ref{fig:video_mnist} one example of each class is illustrated.

\begin{table}
\caption{Class-specific affine transformation applied on the original MNIST digits.}
\label{tab:class_transformations}
\centering
\begin{tabular}{|c|l|c|} \hline
\textbf{digit} & \textbf{transformation description} & \textbf{parameter(s)} \\ \hline
0              & moves horizontally                                                 & $t_x$        \\
1              & moves vertically                                                   & $t_y$        \\
2              & scales down and then up                                            & $s$        \\
3              & rotates clockwise                                                  & $r$        \\
4              & rotates counter-clockwise                                          & $r$        \\
5              & moves along a circular path                                                  & $t_x$, $t_y$     \\
6              & scales up while rotating clockwise                                 & $s$, $r$     \\
7              & moves horizontally while rotating counter-clockwise                & $t_x$, $r$     \\
8              & rotates clockwise and then counter-clockwise                       & $r$        \\
9              & random rotation and horizontal+vertical movements                  & $r$, $t_x$, $t_y$ \\
\hline
\end{tabular}
\end{table}

\subsection{Model architectures}
\label{sec:models}
We use the LeNet-5 architecture~\cite{lecun1998gradient} as it is a good starting point for training a model based on a variant of the MNIST dataset. The original 2D convolutions are replaced with regular 3D convolutions and 3DTT convolutions for the LeNet-5-3D and the LeNet-5-3DTTN respectively. The number of filters in each convolutional layer can vary since during experimentation we noticed that we can achieve better performance by either increasing or reducing the number of filters in the convolutional layers for both LeNet-5-3D and LeNet-5-3DTTN. LeNet-5-3D serves as the baseline model.

\subsection{Training and inference}
\label{sec:training}
\subsubsection{Training}
All models are optimized using SGD with mometum of $0.9$. Depending on the model, the starting learning rate value can vary from $1e-8$ to $5e-9$. The models are trained for a total of $100$ epochs where every $10$th epoch the learning rate decreases exponentially if the validation accuracy has not improved. We noticed that $100$ epochs provides a good time-window for the models to converge. Generally a batch size of $20$ is used unless we are training on only $10$ videos, in which case a batch size of $10$ is used. Initialized LeNet-5-3D model weights as well as all fully connected layers follow a Kaiming-Uniform scheme\cite{he2015delving}\footnote{The default setting in PyTorch.}. LeNet-5-3DTTN initializes $K_0$ with weights sampled from a Gaussian distribution\footnote{Experimentally this gave the best results, however there was very little difference between different types of initializations.}. In our main experiments we use a parameterization of $\Theta$ with the following initialization: $s=1$, $r=0$, $t_x=0$ and $t_y=0$. 

\subsubsection{Replication of video selection}
Each model is ran $30$ times (runs) with the same initialization parameters but with different randomly initialized weights for the convolution and fully connected layers. The training data is randomly selected across different runs by using a seed. The seed assures that the same videos are chosen again when we execute the same run with a different model or when we use different initialization parameters. 
This way we can compare only the difference between the model architecture and parameters without confounding our results with video variance. Given that we experiment with very few videos, we make sure that the classes are represented equally in the randomly selected training data. 

\subsubsection{Inference}
\begin{figure}[h!]
    \centering
    \includegraphics[width=1\linewidth]{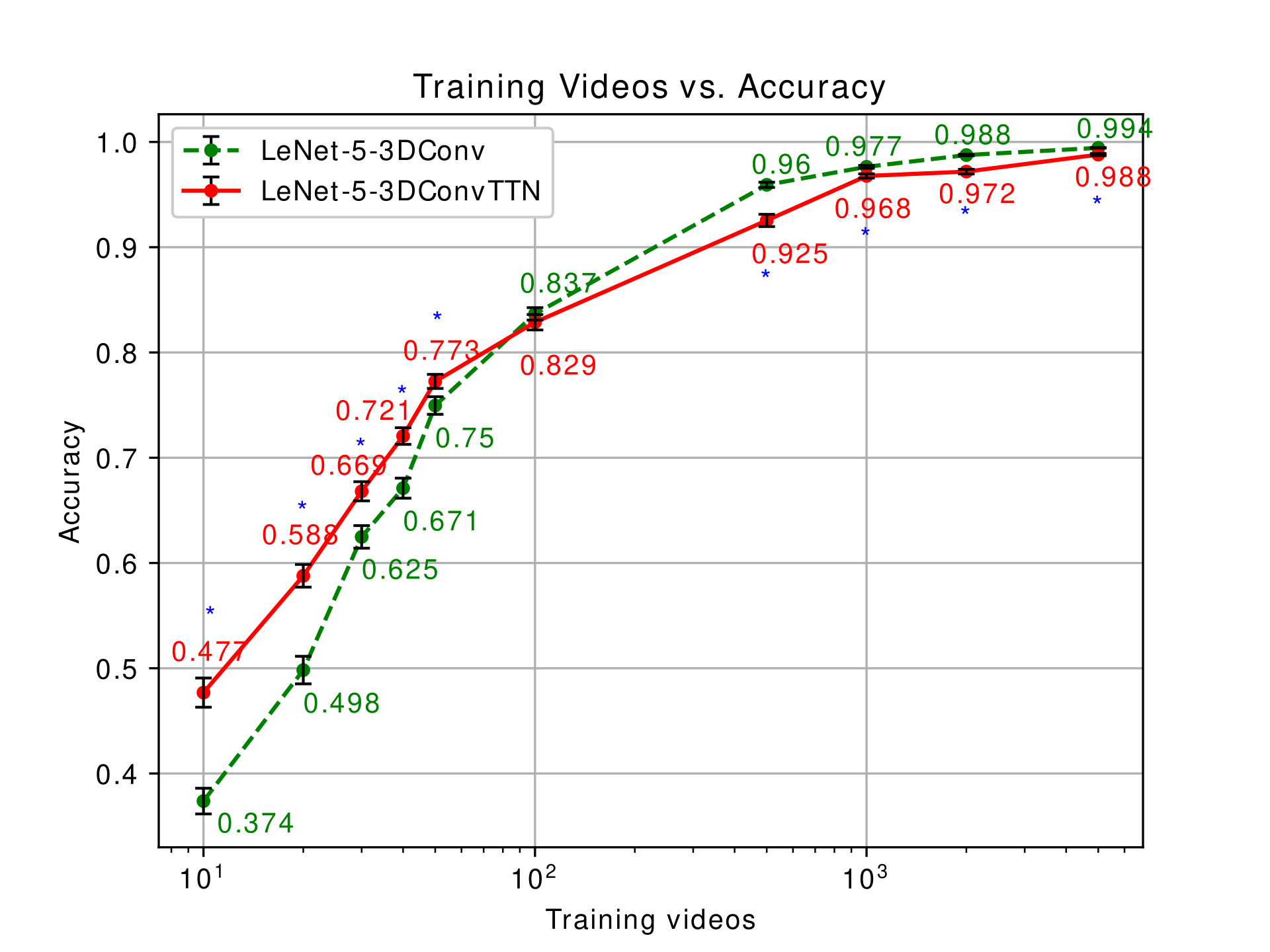}
    \caption{A comparison of the model using regular 3D convolutions with the model using our factorized 3D convolutions on the entire test split of Video-MNIST. An asterisk denotes if the difference between models is significant.}
    \label{fig:comparison_test}
\end{figure}

Model selection is based on the accuracy of the validation split. The 30 models in the run with highest average accuracy are ran against the test split. Each run is essentially the same model using the same hyperparameters but with different randomized weight initializations for the convolutional and fully connected layers. In the end the test results of the 30 models are averaged and the standard error of the results is calculated. The final results can be seen in Figure \ref{fig:comparison_test}. 

\subsection{Setup}
To test if our method can outperform conventional 3D convolutions on very few datapoints we train each separate model on a different number of training videos. This way we can test how efficient our method is.  The total number of videos are varied from low to high: 10, 20, 30, 40, 50, 100, 500, 1000, 2000, 5000. Model selection happens for each of the number of videos separately. The models trained on 10 videos are different from the models trained on 20 videos. After model selection based on the validation split the models from the best run perform inference on the test split.

\begin{table}[h]
\caption{Detailed results LeNet-5-3D vs. LeNet-5-3DTTN on Video-MNIST.}
\centering
\begin{tabular}{|c|c|c|c|c|c|c|}
\hline
\multicolumn{1}{|l|}{} & \multicolumn{3}{c|}{\textbf{LeNet-5-3D}} & \multicolumn{3}{c|}{\textbf{LeNet-5-3DTTN}} \\ \cline{2-7} 
\begin{tabular}[c]{@{}c@{}}train\\ videos\end{tabular} & \begin{tabular}[c]{@{}c@{}}mean\\ accuracy\end{tabular} & \begin{tabular}[c]{@{}c@{}}standard\\ error\end{tabular} & \begin{tabular}[c]{@{}c@{}}model\\ parameters\end{tabular} & \begin{tabular}[c]{@{}c@{}}mean\\ accuracy\end{tabular} & \begin{tabular}[c]{@{}c@{}}standard\\ error\end{tabular} & \begin{tabular}[c]{@{}c@{}}model\\ parameters\end{tabular} \\ \hline
10 & 0.374 & 0.0122 & 375668 & \textbf{0.477} & 0.0139 & 444966 \\
20 & 0.498 & 0.0131 & 496430 & \textbf{0.588} & 0.0108 & 348612 \\
30 & 0.625 & 0.0107 & 496430 & \textbf{0.668} & 0.0091 & 444966 \\
40 & 0.671 & 0.0095 & 496430 & \textbf{0.721} & 0.0079 & 348612 \\
50 & 0.750 & 0.0083 & 496430 & \textbf{0.773} & 0.0067 & 444966 \\
100 & \textbf{0.837} & 0.0059 & 496430 & 0.829 & 0.0073 & 348612 \\
500 & \textbf{0.960} & 0.0025 & 496430 & 0.925 & 0.0058 & 222940 \\
1000 & \textbf{0.976} & 0.0014 & 496430 & 0.968 & 0.0020 & 254058 \\
2000 & \textbf{0.988} & 0.0007 & 496430 & 0.972 & 0.0022 & 222940 \\
5000 & \textbf{0.994} & 0.0003 & 496430 & 0.988 & 0.0010 & 222940 \\ \hline
\end{tabular}
\end{table}

\section{Results}
In Table~\ref{fig:comparison_test} and Figure~\ref{fig:comparison_test} we can see that the that our method outperforms the conventional 3D convolution significantly in the low data regime. However, when we have ample training data the conventional 3D convolution outperforms our method, as is to be expected. It is worth mentioning that, in general, our method uses fewer parameters and still achieves reasonable results in all settings. 

\section{Conclusion}
We propose a novel factorization method for 3D convolutional kernels. Our method factorizes the 3D kernel along the temporal dimension and provides a way to learn the 3D kernel through transformations of a 2D kernel, thereby greatly reducing the number of parameters needed. We demonstrate that our method significantly outperforms the conventional 3D convolution in the low data regime ($10$ to $50$ training videos), yielding 0.58 vs. 0.65 on average for the LeNet-5-3D and the LeNet-5-3DTTN respectively. Additionally our model achieves competitive results in the high data regime ($>100$), with 0.95 vs. 0.94 on average for the LeNet-5-3D and the LeNet-5-3DTTN respectively, using up to $45\%$ fewer parameters. Hence, 3DTTNs provide a useful building block when estimating models for video processing in the low data regime. In future work, we will explore in which real-world problem settings 3DTTNs outperform their nonfactorized counterparts.

\newpage
\bibliographystyle{splncs04}
\bibliography{ref}

\begin{thebibliography}{10}
\providecommand{\url}[1]{\texttt{#1}}
\providecommand{\urlprefix}{URL }
\providecommand{\doi}[1]{https://doi.org/#1}

\bibitem{baccouche2011sequential}
Baccouche, M., Mamalet, F., Wolf, C., Garcia, C., Baskurt, A.: Sequential deep
  learning for human action recognition. In: International workshop on human
  behavior understanding. pp. 29--39. Springer (2011)

\bibitem{goodfellow2016deep}
Goodfellow, I., Bengio, Y., Courville, A., Bengio, Y.: Deep learning, vol.~1.
  MIT Press (2016)

\bibitem{he2015delving}
He, K., Zhang, X., Ren, S., Sun, J.: Delving deep into rectifiers: Surpassing
  human-level performance on imagenet classification. In: Proceedings of the
  IEEE international conference on computer vision. pp. 1026--1034 (2015)

\bibitem{jaderberg2015spatial}
Jaderberg, M., Simonyan, K., Zisserman, A., et~al.: Spatial transformer
  networks. In: Advances in neural information processing systems. pp.
  2017--2025 (2015)

\bibitem{ji20123d}
Ji, S., Xu, W., Yang, M., Yu, K.: 3d convolutional neural networks for human
  action recognition. IEEE transactions on pattern analysis and machine
  intelligence  \textbf{35}(1),  221--231 (2012)

\bibitem{karpathy2014large}
Karpathy, A., Toderici, G., Shetty, S., Leung, T., Sukthankar, R., Fei-Fei, L.:
  Large-scale video classification with convolutional neural networks. In:
  Proceedings of the IEEE conference on Computer Vision and Pattern
  Recognition. pp. 1725--1732 (2014)

\bibitem{lea2016segmental}
Lea, C., Reiter, A., Vidal, R., Hager, G.D.: Segmental spatiotemporal cnns for
  fine-grained action segmentation. In: European Conference on Computer Vision.
  pp. 36--52. Springer (2016)

\bibitem{lea2016temporal}
Lea, C., Vidal, R., Reiter, A., Hager, G.D.: Temporal convolutional networks: A
  unified approach to action segmentation. In: European Conference on Computer
  Vision. pp. 47--54. Springer (2016)

\bibitem{lecun1989generalization}
Lecun, Y.: Generalization and network design strategies. In: Connectionism in
  perspective. Elsevier (1989)

\bibitem{lecun1998gradient}
LeCun, Y., Bottou, L., Bengio, Y., Haffner, P., et~al.: Gradient-based learning
  applied to document recognition. Proceedings of the IEEE  \textbf{86}(11),
  2278--2324 (1998)

\bibitem{qiu2017learning}
Qiu, Z., Yao, T., Mei, T.: Learning spatio-temporal representation with
  pseudo-3d residual networks. In: proceedings of the IEEE International
  Conference on Computer Vision. pp. 5533--5541 (2017)

\bibitem{sun2015human}
Sun, L., Jia, K., Yeung, D.Y., Shi, B.E.: Human action recognition using
  factorized spatio-temporal convolutional networks. In: Proceedings of the
  IEEE international conference on computer vision. pp. 4597--4605 (2015)

\bibitem{tran2018closer}
Tran, D., Wang, H., Torresani, L., Ray, J., LeCun, Y., Paluri, M.: A closer
  look at spatiotemporal convolutions for action recognition. In: Proceedings
  of the IEEE conference on Computer Vision and Pattern Recognition. pp.
  6450--6459 (2018)

\bibitem{varol2017long}
Varol, G., Laptev, I., Schmid, C.: Long-term temporal convolutions for action
  recognition. IEEE transactions on pattern analysis and machine intelligence
  \textbf{40}(6),  1510--1517 (2017)

\bibitem{yue2015beyond}
Yue-Hei~Ng, J., Hausknecht, M., Vijayanarasimhan, S., Vinyals, O., Monga, R.,
  Toderici, G.: Beyond short snippets: Deep networks for video classification.
  In: Proceedings of the IEEE conference on computer vision and pattern
  recognition. pp. 4694--4702 (2015)

\end{thebibliography}
\end{document}